\documentclass[runningheads]{llncs}
\usepackage{comment}
\usepackage{multirow}
\usepackage{graphicx}
\begin{document}
\title{Using Natural Language Processing to Understand Reasons and Motivators Behind Customer Calls in Financial Domain}
\titlerunning{Using NLP to Understand Customer Call Reasons and Motivators}
%
\author{Ankit Patil\thanks{Equal Contribution}\orcidID{0000-0001-8970-1839} \and
Ankush Chopra\mbox{*}\orcidID{0000-0002-9970-8038} \and
Sohom Ghosh\mbox{*}\orcidID{0000-0002-4113-0958} \and
Vamshi Vadla\mbox{*}\orcidID{0000-0002-0966-7552}
}
\authorrunning{Patil et al.}
%
\institute{Fidelity Investments, AI CoE, Bengaluru, India
\email{\{ankitpatil4,ankush01729,sohom1ghosh,vamsi.lg\}@gmail.com}}
\maketitle              
\begin{abstract}
In this era of abundant digital information, customer satisfaction has become one of the prominent factors in the success of any business. Customers want a one-click solution for almost everything. They tend to get unsatisfied if they have to call about something which they could have done online. Moreover, incoming calls are a high-cost component for any business. Thus, it is essential to develop a framework capable of mining the reasons and motivators behind customer calls. This paper proposes two models. Firstly, an attention-based stacked bi-directional Long Short Term Memory Network followed by Hierarchical Clustering for extracting these reasons from transcripts of inbound calls. Secondly, a set of ensemble models based on probabilities from Support Vector Machines and Logistic Regression. It is capable of detecting factors that led to these calls. Extensive evaluation proves the effectiveness of these models.

\end{abstract}

\section{Introduction}
Phone calls are one of the main channels through which customers interact with organizations. Customers call either seeking answers to their queries or for getting a service request fulfilled. For contacting customer care, typically customers need to select the right options in the Interactive Voice Response (IVR) after going through the menu and there is generally a caller queue resulting in a hold time. Hence, phone calls tend to be a slower medium of communication by design compared to mediums like chat, virtual assistants (VA) and search.

In most cases, phone interactions result in suboptimal experience from both organization and customers’ points of view, due to the time it takes to get the information. It also results in higher operational cost of call centres.

There has been a push for digitization of services to enable and empower customers towards self-service. This will enable faster access to information, efficient service request resolution for the customers while reducing the cost of service for the organizations.

For effective digitization of services, it becomes essential to understand the reasons behind the customer calls and the gaps in the current digital experience. In this paper, we propose a framework to understand the reasons and motivators of these calls using Machine Learning and Natural Language Processing (NLP) techniques. These calls are inbounded to Fidelity Investments\footnote{ https://www.fidelity.com/}  and are specific to the financial domain.  

  
\subsection{Understanding Call Reasons}
We summarize the call conversation into customer intents of length up to 6 words. By performing the abstractive summarization on the stored transcripts of these calls. This provides us with the reason behind the call. We have used the Long Short Term Memory network \cite{lstm} with attention \cite{vaswani2017attention} to generate the summary intents. 

\subsection{Understanding Call Motivators}
In this work, we have researched solutions about ``why the customer had to call in": is it because they are not tech-savvy, or they were not able to find the information they are looking for or something for which there is no digital solution available at present. Innovations in this space are a win-win for both the customer (information at fingertip) and for business (cost-saving by reducing inbound calls). 

\subsection{Challenges}
Data scarcity is the biggest challenge when coming up with a solution for the stated problem. A dataset where each of the calls has been tagged with the primary reason and motivation of the call would be the ideal dataset for coming up with solutions. Multiple transfers during the call make it difficult to always decipher the actual reason for the call. Also, there can be many latent motivations for the call, as this is very subjective at times. E.g. \textit{``I need help with the website"} or \textit{``I have attempted to do this online but was unable"}. The former says the customer was not aware of what to do on the website, but the latter informs us that the customer had tried it and was unable to do it online. This may be because of some misleading information or some issue with the pertaining account.

We made use of customer-representative summaries that were available for a fraction of calls to generate the data needed for training the model capable of detecting reasons behind calls. For call motivators, we took the help of the domain experts and representatives who listened to these calls and assigned one label (primary motivator) to every call. Coming up with the labels was an iterative process. After few iterations, we decided to focus on 11 call motivators which were of major interest to the stakeholders. As this list is not exhaustive, we introduced an ``other” label as one broad category which can be mined later for any more categories.\\
\textbf{Contributions} Our contributions are as follows: i) We developed a summarization model and a subsequent clustering model to understand reasons relating to these calls ii) We developed a classification model to comprehend the factors which led to these calls

Combining findings from both models gave us a clear view of the avenues where digitization can be done or improved. We will be describing the dataset in the next section (\ref{sec:data}) and followed by the solution methodology (\ref{sec:method}). We will shed some light on the experimentation and results (\ref{sec:exp}) after the methodology section. We will discuss our findings (\ref{sec:dis}) and conclude (\ref{sec:con}) the paper by narrating the future enhancements to this framework. 

\subsection{Problem Statement}
Our problem statement is two folds. Given a set C = \{c\textsubscript{1}, c\textsubscript{2}, c\textsubscript{3} ... c\textsubscript{n}\} of calls specific to financial domain, we need to i) Identify call reasons and cluster them. Call reasons are an open set of customer issues (like ``adding a bank account",``password reset", ``transfer of assets" and so on) ii) Classify these calls into 12 pre-defined categories of motivators mentioned in Table \ref{tab:reasonsdata}.

\begin{table}
\centering
\caption{Call Motivators and their corresponding occurrence}
\label{tab:reasonsdata}
\begin{tabular}{|l|l|}
\hline
\textbf{Motivators}                                & \textbf{Occurrence(\%)} \\ \hline
 M1        & 22.94                   \\ \hline
M2 & 1.21                    \\ \hline
M3             & 2.47                    \\ \hline
M4           & 1.02                    \\ \hline
M5                            & 3.91                    \\ \hline
M6            & 18.44                   \\ \hline
M7                                   & 6.68                    \\ \hline
M8         & 13.90                   \\ \hline
M9                     & 1.67                    \\ \hline
M10           & 14.09                   \\ \hline
M11           & 7.42                    \\ \hline
Other                                                   & 6.24                    \\ \hline

\end{tabular}
\end{table}

\section{Related Works}
Some of the conventional extractive text summarization techniques include Text-Rank \cite{mihalcea-tarau-2004-textrank} (a graph-based ranking model) and LexRank \cite{lexrank}. Text-Rank is an unsupervised method for extracting keywords and sentences. LexRank is a graph centrality based model to score sentences. \\
Some of the newer text summarization techniques include genetic algorithm based approach MUSE \cite{litvak-etal-2010-new} and linear programming based approach POLY \cite{litvak-vanetik-2013-mining}. Shi et al. in their survey paper \cite{shi2020neural} described various neural network based abstractive summarization processes. They further developed and open-sourced an abstractive summarization library named Neural Abstractive Text Summarizer (NATS) toolkit. Gupta et al. in their paper \cite{Gupta2010ASO} surveyed several extractive summarization techniques. In the paper \cite{verma2007semantic}, Verma et al. described an ontology-based summarization model tuned specifically for Medical documents. They used WordNet and Unified Medical Language System to revise keywords for ranking sentences. 
The Financial Narrative Summarization Shared Task (FNS 2020) \cite{el-haj-etal-2020-financial} was organised by El-Hai et al. This task was about summarizing annual reports of organizations listed in London Stock Exchange, UK. They discussed the efforts of 9 teams. The best performing team SUMSUM comprising Zheng et al. \cite{zheng-etal-2020-sumsum} parsed these financial reports into sections based on Table of Contents (ToC) and applied BERT \cite{devlin2018bert} to each of them. They achieved a F1 score of 0.306. Singh \cite{singh-2020-point} ranked second by achieving a F1 score of 0.289 using Pointer Network \cite{see-etal-2017-get}  and T5 – Text to Text Transfer Transformer \cite{T5}. Azzi et al. of team FORTIA \cite{ait-azzi-kang-2020-extractive} ranked third. They used rule-based approaches to extract ToC. They also used a Convoluted Neural Network-based binary classifier to identify candidate summaries. Their F1 score was 0.274. 
A goal guided summarization \cite{goalgguided} model had been described by Agrawal et al. in their paper \cite{goalgguided}. They
 narrated how they summarized annual reports of organizations using hierarchal neural network models to infer whether to buy or sell a stock.\\
 So, far we discussed various research works on financial text summarization. Now, we will discuss prior works related to financial text classification. Ciravegna et. al \cite{facile}
developed a financial text classification system called FACILE. This system was flexible and it worked for corpora of 4 different languages namely Italian, English, German and Spanish. Zhao et al. \cite{9102263} described how they used partial information to classify Chinese Financial News. Yang et al. in their paper \cite{yang2020generating} discussed how much explainable transformer-based financial text classification models are. Various pre-processing techniques for classifying financial texts had been described by Sun et al. in their paper \cite{preprocess}.\\
It is quite interesting to note that while a lot of work has been in the area of financial text processing, only a few of them (\cite{hmmextractivesummarization} and \cite{chu-carroll-carpenter-1999-vector}) tried to understand the intent behind inbound calls specific to the financial domain. Moreover, none of them performed abstracting summarization followed by hierarchical clustering and classification of customer calls related to the financial domain to analyse the reasons and motivators behind these calls.


\section{Data}
\label{sec:data}
\subsection{Data used for training Call Reason Models}
We first pre-process and clean call transcripts. We remove the system noise, the transcription-induced noise, and masked tokens. We then remove the pleasantries and other non-informative phrases. We also perform case normalization and contraction replacement. Customer-care representatives summarize the conversations they had during a call and notes them down at the end of the call. These summary notes are only available for a fraction of calls (\~20\%). We call these summaries repnotes. Next, we clean repnotes by performing steps like lower case conversion, removal of masked tokens, replacement of contraction and non-informative phrases. Details of the pre-processing steps can be found in Table \ref{tab:preprocessing}.

\begin{table}
\centering
\caption{Pre-processing details for Call Reason Models}
\label{tab:preprocessing}
\begin{tabular}{|l|c|c|}
\hline
\textbf{Preprocessing (Normalization) Step} &
  \multicolumn{1}{l|}{\textbf{Call Transcripts}} &
  \multicolumn{1}{l|}{\textbf{RepNotes}} \\ \hline
\begin{tabular}[c]{@{}l@{}}Call transcript data contain system\\ messages like \textit{``party has left the session"}\end{tabular} &
  Yes &
  No \\ \hline
Lower case conversion                        & Yes & Yes \\ \hline
Replacing the masked tokens                  & Yes & Yes \\ \hline
Removal of non-vocalized noise transcription & Yes &     \\ \hline
Contraction replacement                      & Yes & Yes \\ \hline
\begin{tabular}[c]{@{}l@{}}Removal of non-informative prefixes \\ like ``customer contacted" or ``customer asked"\end{tabular} &
  No &
  Yes \\ \hline
\end{tabular}
\end{table}

We looked at the length distribution of cleaned call transcripts and repnotes. 90\% of the call transcripts were of 425 words or less, with a long tail for the remaining 10\%. Repnotes with a length of 6 or less constitute 35\% of the entire repnotes data. 40\% of them are of length between 7 to 17 and the rest have 18 words or more. Since we are trying to generate concise customer intents from the calls, we decided to use the repnotes  which were in the first bucket. We divide the data into the train (70\%) and validation (30\%) set. We considered all the calls with less than or equal to 425 tokens in transcripts where repnotes of length 6 or less were present for modelling. This was close to 1.9 million interactions. 

\subsection{Data used for training Call Motivator Models}
\subsubsection{Data Labeling Exercise} 
Mining motivators from customer calls is a hard task mainly due to the subjectivity associated with it. A labelling exercise was designed to know what kind of different call motivators can be present in the financial domain. It was an iterative process executed by a variety of customer representatives. Finally, 11 categories are selected but those did not cover 100\% of the calls. Thus, an “Other” label was introduced. In all, we have 12 categories of business interests (primary call motivators) distributed over ~6000 inbound calls. The data set is highly imbalanced, with the most frequent category ``\textit{Attempted to do something online but was unable to}" had an incidence of ~23\% and most infrequent category ``\textit{Following up a communication}" had incidence of only ~1\%. The details mentioned in Table \ref{tab:reasonsdata}. Due to internal security reasons, we mask these motivators as M1, M2 ... M11. We used 70\% of this data for training and 30\% for validation.

\subsubsection{Data Discovery}

With labelled data available, this becomes a text classification problem for us. The call interaction was available to us in the form of transcripted text. These transcripts (noisy at times) were not enough. So, we have explored other data streams around an inbound customer call. 

Most of the customers leave trails of data before and after the call: 
\begin{enumerate}
\item     Before calling, customers try to find information online, so they visit the website (clickstream data). 

\item     When they call, they need to state their need (whisper) to help us route the call. 

\item     During the call, they might browse through the website (clickstream). 

\item     After the call, we have the transcript of the call.

\item     The representative writes a note/summary of the call for future references. 

\item     We also have a model, which explains the reason for a call and further generates a summary of the call interaction.
\end{enumerate}

\section{Methodology}
\label{sec:method}
This work is divided into two major parts, identifying the reasons behind the calls and classifying the call motivators. We initiate by describing the first of the two major parts of the framework. 
\subsection{Understanding Call Reasons}
Call transcripts are stored at the utterance level, where every entry is the transcription of what the agent or customer has said. Calls have an introduction, pleasantries, and informal conversation messages along with the actual business conversation. This makes transcripts lengthier where customer intent is hidden somewhere in the body of the conversation. To be able to understand the reason behind the call, we summarize the call into short customer intent. We use the transcript and human-generated summaries to create the call reason model.  

Customer calls are transcribed and stored in the form of text. We use these transcripts with repnotes. Reps inadvertently end up introducing a lot of variations and subjectivity into the repnotes even if they mean the same thing. For example, 
\textit{``customer contacted to get account reset"} and \textit{``customer asked for the help with password reset although worded differently"}, mean the same.

We have used transcribed calls and the repnotes as the training data for the intent generation model. Call transcripts are used as input to the model and repnote are used as summary intent or the reason behind the call. Since we are looking for short customer intents, we limit our data to only inbound calls where repnotes was written with 6 words or less. 

Repnotes are abstractive summarizations of the conversation that happened in customer calls. We use LSTM \cite{lstm} architecture with attention \cite{vaswani2017attention} to model this. Attention particularly helps here to handle the transcripts well which otherwise be difficult due to their length. It is a 2 layered stacked bi-LSTM sequence-to-sequence with attention architecture. We have not used any pre-trained embedding due to the nature of the underlying data. Instead, we are learning the embedding while training for this task. 

\begin{figure}[ht]
 \centering
\includegraphics[width=10cm]{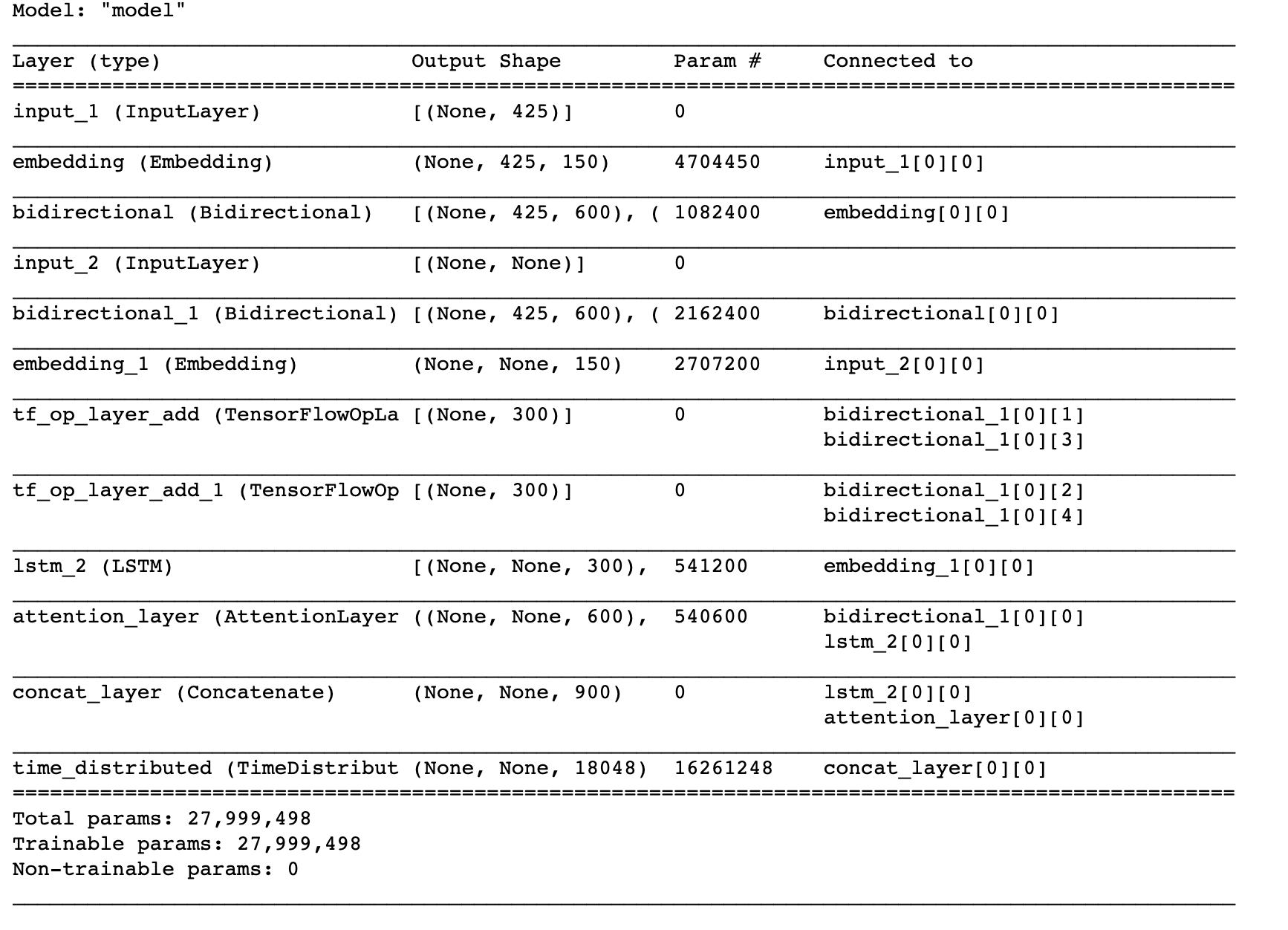}
\caption{Hyper-parameters of the final Attention based stacked Bi-LSTM Model}
\label{fig:summhhyper}
\end{figure}

The Model hit the early stopping with a validation categorical cross-entropy loss of 1.884 at the 28\textsuperscript{th} epoch. This model is used to generate the customer intent behind all incoming calls and chat transcripts irrespective of whether they have repnotes or not. These generated repnotes look much more like short customer intents and explain the reason behind the calls. The model summary has been mentioned in Figure \ref{fig:summhhyper}. Other details include: Latent dimension = 300, Embedding dimension = 150, size of vocabulary for repnote = 18048, size of vocabulary for input transcript = 31363, optimizer = rmsprop and loss = sparse categorical crossentropy.

These generated intents are sufficient to explain the reasons behind the calls. Every year our organization, Fidelity Investments receives millions of customer calls. A framework to digest this data easily is immensely useful. We added a layer of clustering to make this data more actionable.  We take all the generated customer intents and apply some preprocessing and normalization steps. We replace acronyms with their definition, remove repetitive words, and remove any customer-specific information which may have been captured in the generated summary. Further, we convert these cleaned customer intents to their equivalent mathematical representation or phrase embeddings. We used sentence transformers \cite{sbert} to come up with phrase embeddings while using the RoBERTa \cite{liu2019roberta} embeddings at the back. Finally, we use  Agglomerative clustering as mentioned in \cite{chopra2021datafnnp} to group these call reasons into homogeneous buckets.

\subsection{Understanding Call Motivators}

We have approached building a classifier in 2 ways, multi-class classification and 1-vs-all classification. Multi-class approach had not resulted in a good model, mainly due to the imbalanced dataset. So, we had dropped it at the early stage of this exercise after trying techniques like over-sampling and under-sampling. 
We need to build models for 12 motivators as mentioned in the earlier section. We have designed a framework that consists of a variety of techniques at each step and finally selects a model based on the chosen evaluation metric. The process followed to get to 1-vs-all classification models has 5 components as described below:

\begin{enumerate}

    \item     \textbf{Input Features:} Count and Term Frequency Inverse Document Frequency (TF-IDF) based vectorizers on unigrams and unigrams-bigrams. 

    \item    \textbf{Feature Reduction:} PCA (100 dimensions), LDA (100 topics) 

    \item    \textbf{Classification Algorithms:} Regularized Logistic Regressions, Support Vector Machine \cite{svmvapnik1995support}, Gradient Boosting Machines (GBM) \cite{gbm}. 

    \item  \textbf{Hyperparameter Tuning:} Using hyperparameter search library Optuna\footnote{https://optuna.org/ (Accessed on 15\textsuperscript{th} June 2021)} and Grid-search. 

    \item  \textbf{Model selection:} Based on high-performance evaluation metric. 
\end{enumerate} 
We have applied the same process to the different data sources and combined some of them. The final data sources we have used: 
\begin{enumerate}
    \item  \textbf{Call transcript:} We have built models separately using transcripts. 

    \item   \textbf{Whisper and Summary:} whisper is stated by the customer and summary (build using rep-notes) is from the call transcripts. These 2 datasets complement each other, so it is logical to combine them. These resulted in a good performance on some of the call motivators. 

    \item  \textbf{Clickstream data:} We have built models using custom features like, session length, page tags, dwell time etc.
    \item After getting models from each of these datasets, the scores are used as features to build ensemble models for each call motivator. The ensembles have given good lift in precision as well as recall.
\end{enumerate} 
The best models for each motivator had different components, some models are built using TF-IDF followed by PCA as input features to Logistic Regression, some had Binary Counts followed by LDA as input features to SVM. The final model for 8 of the 12 motivators is an ensemble model built on probability scores from different models. The rest of the models are call transcript-based model with different combination of the above techniques. 
\section{Experimental Setup and Results}
\label{sec:exp}
In this section, we narrate the experiments we performed and their results. We performed these experimentations on 
a Nvidia DGX GPU cluster (having 160,000 CUDA Cores and over 20,000 Tensor Cores) and an internal cloud instance (having 64GB RAM and 16 processing units).

\subsection{Experiments related to Call Reason Models}
We first tried a simple sequence-to-sequence (s2s) architecture with LSTM. The simple s2s model did not capture the long-term dependencies very well, and since the call transcripts were too long it didn’t do well for this task. Next, we tried the s2s model with attention which significantly helped in improving the model's performance (i.e. in minimising the categorical cross-entropy loss) in the validation set. We experimented with the number of layers in LSTM. We tried 1, 2 and 3 layers. We noticed an improvement in performance while using two layers compared to one layer, but there was no significant performance gain when we tried 3 layers. We also experimented with both unidirectional and bidirectional LSTMs, where bidirectional LSTMs performed better. We were running into exploding gradient issue and applied gradient clipping to avoid it.  

We also tried pre-trained transformer-based fine-tuning using BART \cite{BART} and T5 \cite{T5} but they surprisingly did not do very well. On closer look, it is due to the poorly transcribed data, which did not have well-formed sentences and phrases. Looking at the performance of the above models, we used s2s with the attention model for generating the interaction intent or repnotes.

\subsubsection{Model Performance}
Since manually validating predictions for the whole data is difficult, we randomly sample 2000 hold-out instances consisting of themes and their corresponding repnotes for evaluating the performance of the model. We also evaluate them automatically. For automatic evaluations we used Rogue-1 and Rogue-L \cite{lin-2004-rouge}. These results have been presented in Table \ref{tab:summarization-results}. It reveals that stacked bi-directional LSTM gives the best precision in terms of Rouge score \cite{lin-2004-rouge}. For manual evaluation, we used the tagged instances and compared them with their corresponding transcripts. We found that in 76.70\% of cases the outputs were acceptable.

\begin{table}
\centering
\caption{Summarization Model Results with highest precisions marked in \textbf{bold}}
\label{tab:summarization-results}
\begin{tabular}{|l|l|l|l|l|}
\hline
               & \multicolumn{2}{l|}{\textbf{Rouge-1}} & \multicolumn{2}{l|}{\textbf{Rouge-L}} \\ \hline
\textbf{Model} & \textbf{Precision}                      & \textbf{Recall}            & \textbf{Precision}                      & \textbf{Recall}            \\ \hline
LSTM           & 0.32                   & 0.16         & 0.29                   & 0.15         \\ \hline
Bi-LSTM         & 0.36                   & 0.17         & 0.33                   & 0.16         \\ \hline
stacked Bi-LSTM & \textbf{0.41}          & 0.13         & \textbf{0.40}          & 0.13         \\ \hline
BART           & 0.38                   & 0.14         & 0.36                   & 0.14         \\ \hline
T5             & 0.38                   & 0.16         & 0.37                   & 0.15         \\ \hline
\end{tabular}
\end{table}

\subsection{Experimentation related to Call Motivator Models}

Due to the nature of call motivators, a one-size-fits-all technique did not work. Hence, we have experimented with a variety of traditional text-classification techniques as mentioned in Table \ref{tab:modelready}. 

\subsubsection{Evaluation Metric}

From our experience, we know it is always better to define the metric based on usage of the model by the business. In our case, the predicted motivations for the call will drive the customer-centric solutions. This downstream process becomes costly as it will involve a lot of human resources and capital. If the model is not precise enough, it will lead towards a different solution which might not solve the customer’s problem and all this effort might be wasted. 

This helped us to decide precision as the metric for evaluation as the cost associated with the wrong prediction is very high. We also ensure that the  Recall is over a given threshold. Also, we have enough call data to compensate for the loss of recall. The results on the validation set have been presented in Table \ref{tab:modelperfromance} (P and R denote precision and recall respectively).

\begin{table}
\centering
\caption{Model ready pipeline}
\label{tab:modelready}
\begin{tabular}{|l|l|c|}
\hline
\textbf{Module}          & \textbf{Techniques}                        & \multicolumn{1}{l|}{\textbf{RepNotes}} \\ \hline
Text Preprocessing       & Stop-word removal, Lemmatization, Stemming & No                                     \\ \hline
Feature Extraction       & Count, TF-IDF, BERT Embeddings             & Yes                                    \\ \hline
Dimensionality Reduction & PCA, LDA (Topic Models)                    & Yes                                    \\ \hline
\end{tabular}
\end{table}

\subsubsection{Experimentation with Call Transcript Data (CT)}
Call transcripts are the most important data stream for predicting the call motivator as it directly comes from the source of the call (call audio). But transcripts can be noisy at times either because of the performance limitation of the speech-to-text system or due to the background noise at the side of customer. This directly impacts the performance of the models built on transcripts. 
After text pre-processing, we have obtained count and TF-IDF vectorizer and used them as features for the model. We also experimented with Logistic Regression, SVM, GBM, AdaBoost \cite{adaboost}. These models had very high variance, mainly due to the very high number of features. Alternatively, we have used PCA and LDA (which were trained on 1 year worth of call-transcripts) to reduce the feature space and later helps us capturing the semantic similarity. These updated models had a comparable performance on the train and validation sets.

\subsubsection{Experimentation with Whisper, Call Summaries \& Rep-Notes Data (WSR)}
Through whisper, we capture direct customer intent and by adding summary/rep-notes with it we can capture what has been discussed in the call. For example, a customer may state his need as \textit{``pin password reset"} and summary data has \textit{``customer tried to pin/password reset but the account was blocked"}. Both gave us good predictors for \textit{``Attempted to do something online but was unable"}. We have observed a good lift on some of the motivators compared to transcript based models. . 

\subsubsection{Experimentation with Clickstream Data (CS)} 
We have also extracted users’ digital footprint of the clicks they had on Fidelity Investment’s website. Again, using the web click-stream is strongly motivated by our hypothesis that some customers call our customer care if they are themselves unable to do what they want on the website. So, it is fair to assume that we can find signals of call motivation from the click footprint they have left on the website right before the call. For example, it could be as complex as them having tried and failed to place a trade or as simple as them failing in changing their password. As we see here, there could be a spectrum of digital difficulty that could lead a customer to place a call to Fidelity’s customer care. We are motivated to identify the motivators to push at least these digitally easy tasks towards self-service through the site. 
From our clickstream data, we extracted user sessions which constituted the sequence of user clicks. We also extracted some other click features such as dwell time, session length, product information of the pages the customer has viewed, and user agent attributes such as device information, OS information, browser information etc. These features had been extracted for various time windows of different lengths. Based on this we chose 3-time windows (1 day \& 1 hour before the call and 1 hour after the call) for which we pulled all the below features which include both features from the source and engineered features. The list of features used from the source includes - total dwell time, page names, purposes which the web pages are about, products listed in them, channel from which users have come and search phrases. The list of engineered features include - pre/post/during session times around the call, average of all session lengths of the call, average number of clicks in the sessions of the call, average number of repetitive click instances in the session, the maximum number of repetitive clicks in the session, number of articles viewed, attribute derived from users (such as device, browser, OS etc.), whether virtual agent/chat has been used and whether any page error has been encountered.

\subsubsection{Creating ensemble from independent data-based models (EM)}

After error analysis on predictions of these models, we have observed that these models have some diverse predictions. For the same motivator, where one model failed, other model(s) were driving the decision. These led us towards building ensemble models which take in the predicted probabilities from the individual models as features. We have built logistic regression models for each of the call motivators, which resulted in a good performance on most of the call motivators.


\begin{table}
\centering
\caption{Precision(P) and Recall (R) on the validation set.}
\label{tab:modelperfromance}
\begin{tabular}{|l|c|c|c|c|c|c|c|c|}
\hline
\multicolumn{1}{|c|}{\textbf{Models}} &
  \multicolumn{2}{c|}{\textbf{CT}} &
  \multicolumn{2}{c|}{\textbf{WSR}} &
  \multicolumn{2}{c|}{\textbf{CS}} &
  \multicolumn{2}{c|}{\textbf{EM}} \\ \hline
\multicolumn{1}{|c|}{\textbf{Motivators}} &
  \textbf{P} &
  \textbf{R} &
  \textbf{P} &
  \textbf{R} &
  \textbf{P} &
  \textbf{R} &
  \textbf{P} &
  \textbf{R}\\ \hline
M1        & .70 & .27 & .57 & .31 & .39 & .77 & \textbf{.86} & \textbf{.28} \\ \hline
M2  & \textbf{.07}  & \textbf{.62} & .03  & .45 & .05  & .18 & \textbf{.07} & \textbf{.62} \\ \hline
M3            & .19 & .33 & .13 & .18 & .04 & .09 & \textbf{.88} & \textbf{.25} \\ \hline
M4                   & .07  & .46 & .43  & .23 & .00 & .00 & \textbf{.43} & \textbf{.23} \\ \hline
M5                          & .42 & .31 & .08 & .66 & .16  & .18 & \textbf{.92} & \textbf{.25} \\ \hline
M6             & .83 & .15 & .24 & .61 & .26 & .28 & \textbf{.83} & \textbf{.19} \\ \hline
M7                                   & \textbf{.86} & \textbf{.14} & .86 & .14 & .40 & .71 & \textbf{.86} & \textbf{.14} \\ \hline
M8          & .66 & .15 & .48 & .08 & .35 & .31  & \textbf{.78} & \textbf{.28} \\ \hline
M9                    & \textbf{.29} & \textbf{.27} & .10 & .31 & .07 & .20 & \textbf{.29} & \textbf{.27} \\ \hline
M10           & .32 & .26 & .29 & .06 & .19 & .18  & \textbf{.77} & \textbf{.16} \\ \hline
M11          & .59 & .11 & .15 & .11 & .20 & .22 & \textbf{.58} & \textbf{.62} \\ \hline
\end{tabular}
\end{table}
 
\section{Discussion}
\label{sec:dis}
Analysing the results of the summarization model, we see that its performance in terms of Rouge scores is not as good as that done using manual validation. This is as per the prior works \cite{liu-liu-2008-correlation} and \cite{ng2015better}. To comprehend this, we inspected some instances manually where the rouge score is much lower than manual evaluation scores. We found that this happened as the themes (generated using abstractive summarization model) were not syntactically similar to the repnotes but their meanings were similar. For example: \textit{``confirmed beneficiaries on file"} and	\textit{``bene verified"}. Moreover, when we further analysed the corresponding call transcripts, we observed that several calls had multiple themes in them. However, the repnotes did not cover all of these themes.\\
With the small data available for detecting motivation of the call, we have experimented with a variety of techniques for feature creation, reduction and classifications using a variety of data sources call transcripts, whisper, clickstream etc. Call transcript-based models have shown good performance over other independent data sources, except for 1 motivator. The ensemble models have outperformed the independent models, as the model was able to capture information from different sources. E.g., for the motivator “Attempted to do something online but was unable”, other than transcript features the clickstream features like “customer sessions on website” was a powerful signal which has increased precision by \~30\%. The ensemble model produced around 4 times increase in precision of motivator M3. Similarly, we have seen significant improvement in precision for other motivators. In our next set of experiments, these high precision models can be used to expand the dataset organically with the help of human annotators and help to improve the recall of the models.  

\section{Conclusion and Future Works}
\label{sec:con}
In this paper, we have introduced an approach that synthesizes the reason for the call with the motivation behind it. It not only extracts call reasons but also determines what the primary motivator was behind the call. Presently, this is being put into production to generate insights that are more informative and interpretable by the business. It drives them to take action by strategising these call reasons and motivators as internal guides to drive business decisions, reduce the calls to action, drive business digitally and boost the performance of specific key performance indicators significantly. The results confirm the benefits of our approach. For further work, the possibility of learning with other structured outputs, increasing the tagged data using a semi-supervised graph algorithm is recommended. Moreover, we would like to extract multiple themes from a call using Beam Search like approach. We would like to monitor the performances of these models and evaluate them using external factors like reduction in call volumes, increment in customer experience index and so on.

%
%
%
 \bibliographystyle{splncs04}
 \bibliography{references.bib}
\end{document}